\documentclass[10pt,twocolumn,letterpaper]{article}

\usepackage{cvpr}
\usepackage{times}
\usepackage{epsfig}
\usepackage{graphicx}
\usepackage{array}
\usepackage{amsmath}
\usepackage{amssymb}
\usepackage{amsthm}

\DeclareMathOperator*{\argmax}{arg\,max}
\usepackage[position=top]{subfig}
\usepackage[export]{adjustbox}
\usepackage{wrapfig}
\usepackage{verbatim}
\usepackage[toc,page]{appendix}

\usepackage{url}
\usepackage{enumitem}
\usepackage{authblk}
\mathchardef\mhyphen="2D
\usepackage{booktabs}
\usepackage{algorithm}
\usepackage[]{algpseudocode}

\usepackage[pagebackref=true,breaklinks=true,letterpaper=true,colorlinks,bookmarks=false]{hyperref}

\cvprfinalcopy 

\def\cvprPaperID{36} 

\begin{document}

\title{Unsupervised Domain Adaptation via Calibrating Uncertainties}



\author[1]{Ligong Han}
\author[2]{Yang Zou}
\author[3]{Ruijiang Gao}
\author[1]{Lezi Wang}
\author[1]{Dimitris Metaxas}
\affil[1]{Department of Computer Science, Rutgers University}
\affil[2]{Department of Electrical and Computer Engineering, Carnegie Mellon University}
\affil[3]{McCombs School of Business, The University of Texas at Austin}
\affil[ ]{\small{ \texttt{l.han@rutgers.edu} \quad \texttt{yzou2@andrew.cmu.edu} \quad \texttt{ruijiang@utexas.edu} \quad \texttt{lw462@cs.rutgers.edu} \quad \texttt{dnm@cs.rutgers.edu} }}


\maketitle
\begin{abstract}
    Unsupervised domain adaptation (UDA) aims at inferring class labels for unlabeled target domain given a related labeled source dataset. Intuitively, a model trained on source domain normally produces higher uncertainties for unseen data. In this work, we build on this assumption and propose to adapt from source to target domain via calibrating their predictive uncertainties. The uncertainty is quantified as the R\'{e}nyi entropy, from which we propose a general R\'{e}nyi entropy regularization (RER) framework. We further employ variational Bayes learning for reliable uncertainty estimation. In addition, calibrating the sample variance of network parameters serves as a plug-in regularizer for training\footnote{This is an extended version of a CVPRW'19 workshop paper with the same title. In the current version the gradient variance regularizer is discussed as a new quantification of uncertainty and a way of regularizing the training dynamics.}. We discuss the theoretical properties of the proposed method and demonstrate its effectiveness on three domain-adaptation tasks.
\end{abstract}

\section{Introduction}
The ability to model uncertainty is important in unsupervised domain adaptation (UDA). For example, self-training~\cite{lee2013pseudo,zou2018unsupervised} often requires the model to reliably estimate the uncertainty of its prediction on target domain in the pseudo-label selection phase. However, traditional deep neural networks (DNN) can easily assign high confidence to a wrong prediction~\cite{gal2016dropout,louizos2017multiplicative}, thus are not able to reliably and quantitatively render the uncertainty given data.

On the one hand, Bayesian neural networks (BNN)~\cite{neal2012bayesian,gal2016dropout,blundell2015weight,kendall2017uncertainties} tackles this problem by taking a Bayesian view of the model training. Instead of obtaining a point estimate of weights, BNN tries to model the distribution over weights. We leverage BNN as a powerful tool for uncertainty estimation. On the other hand, one can estimate the empirical uncertainty of the model by the variance of network parameters, which we call gradient variance regularization (GVR).

Finally, our approach builds on the intuition that a model gives similar uncertainty estimates on the two domains learns to adapt from source to target well. Thus, we propose to directly calibrate the estimated uncertainties between source and target domains during training. This calibration can be considered in three-folds, from which we listed our contributions as follows:
\setlist{nolistsep}
\begin{itemize}
    \item We propose a general framework for unsupervised domain adaptation by calibrating the predictive uncertainty, and discuss its relationship with entropy regularization~\cite{grandvalet2006entropy} and self-training~\cite{lee2013pseudo}.
    \item We introduce variational Bayes neural networks to provide reliable uncertainty estimations.
    \item We propose to calibrate the variance of network parameters as a model-and-objective-agnostic regularization (GVR) on the optimization dynamics.
\end{itemize}

\section{Related Work}
Shannon entropy is commonly used to quantify the uncertainty of a given distribution. Entropy-based UDA has already been proposed in~\cite{vu2018advent}. Unlike~\cite{vu2018advent}, we avoid using adversarial learning which tends to be unstable and hard to train. Also, entropy regularization is proposed in ~\cite{grandvalet2006entropy} for semi-supervised learning and can be directly applied to UDA. However, our framework is more general since the uncertainty is not necessarily to be the Shannon entropy. In fact, we formalize the uncertainty as R\'{e}nyi entropy which is a generalization of Shannon entropy. Many other methods in UDA can be modeled under this framework, for example, self-train~\cite{lee2013pseudo,zou2018unsupervised} can be viewed as minimizing the min-entropy which is a special case of R\'{e}nyi entropy.

As pointed out by~\cite{grandvalet2005semi}, directly optimizing the estimated Shannon entropy given data requires the classifier to be locally-Lipschitz~\cite{miyato2015distributional}. Co-DA~\cite{kumar2018co} and DIRT-T~\cite{shu2018dirt} propose to solve this problem by incorporate the locally-Lipschitz constraint via virtual adversarial training (VAT)~\cite{miyato2015distributional}. 

Another complimentary line of research employs self-ensemble and shows promising results~\cite{french2017self}. Indeed, BNN~\cite{gal2016dropout} performs Bayesian ensembling by nature. This is part of the reason why BNN provides a better uncertainty estimation.

In supervised learning, regularization is proposed to avoid overfitting. Besides weight decay, typical regularization techniques include label smoothing~\cite{goodfellow2016deep,szegedy2016rethinking}, network output regularization~\cite{pereyra2017regularizing}, knowledge distillation~\cite{hinton2015distilling}. We believe our proposed gradient variance regularizer GVR can also be used in supervised settings.

\section{Uncertainty in Deep Neural Networks}
\noindent \textbf{R\'{e}nyi entropy.} For a discrete probability distribution $P=(P_1, \ldots, P_K)$, the R\'{e}nyi entropy~\cite{wiki:renyi} of order $\alpha$ ($\alpha > 0$) is defined as
\begin{align}
    \text{H}_{\alpha}(P)=\frac{1}{1-\alpha}\log(\sum_{k}{P_k^\alpha}).
\end{align}
The limiting value of $\text{H}_\alpha$ when $\alpha \rightarrow 1$ is the {\em Shannon entropy}, and $\alpha \rightarrow \infty$ corresponds to the {\em min-entropy}, $\text{H}_{\infty}(P)=\min_k{-\log(P_k)}=-\log\max_k P_k$. A typical deep neural network for classification usually produces a discrete distribution over possible classes given the input data. Thus, we quantify the predictive {\em uncertainty} by the {\em R\'{e}nyi entropy} on this probability distribution.

\noindent \textbf{Bayesian neural networks.} BNN estimates the posterior over network weights while optimizing the training objective. Given the dataset $\mathcal{D}=\{x^{(i)}, y^{(i)}\}_{i=1}^N$, the output of BNN is denoted as $f(x|w)$ where $x$ is input data and $w$ are the weights. For a classification task, $f$ is the predicted logits and the resulting probability vector is given by a softmax function: $P(y|x,w)=\text{softmax}(f(x|w))$. The predictive distribution over labels given input $x$ is $P(y|x)=\mathbb{E}_{P(w|\mathcal{D})}{P(y|x,w)}$. We define the uncertainty evaluated by BNNs as the entropy $\text{H}_\alpha(P(y|x))$.

We adopt the method from~\cite{kendall2017uncertainties}, where aleatoric and epistemic uncertainties are jointly modeled. In \cite{kendall2017uncertainties}, the logits are assumed to be Gaussian and the reparameterization trick is utilized. The predicted logit is $\hat{f}(x) = \mu_\theta(x) + \sigma_\theta(x) \epsilon$ with $\epsilon \sim N(0, I)$. The final predicted probability vector $P(y|x)$ is approximated by Monte Carlo sampling (with $M$ samples),
\begin{align}
    P(y|x; \theta) = \frac{1}{M}\sum_{m=1}^{M}{\text{softmax}(\hat{f}^{(m)}(x))}.
    \label{eq:pred}
\end{align}

\noindent \textbf{Variational inference.} As estimating the posterior $P(w|\mathcal{D})$ is often intractable~\cite{blundell2015weight,kendall2017uncertainties}, variational inference is commonly adopted, where the posterior of weights is approximated by $Q_\theta(w)\approx P(w|\mathcal{D})$ with parameter $\theta$. Specifically, in supervised learning, $Q_\theta(w)$ is estimated by maximizing the evidence lower bound (ELBO)~\cite{kingma2013auto,gal2016dropout}:
\begin{align}
    ELBO = \underbrace{\mathbb{E}_{Q_\theta(w)}{\log{P(y|x,w)}}}_{(\text{I})}-\underbrace{\text{D}_{KL}(Q_\theta \Vert P(w))}_{(\text{II})},
    \label{eq:elbo}
\end{align}
where $P(w)$ is the prior, and term (I) is the standard cross-entropy loss evaluated at $x$ with parameter $w$. Gal \etal~\cite{gal2015bayesian,gal2016dropout} proposes to view dropout together with weight decay as a Bayesian approximation, where sampling from $Q_\theta$ is equivalent to performing dropout and term (II) becomes $L_2$ regularization (or weight decay) on $\theta$.

\noindent \textbf{Gradient variance.} Rather than finding a variational approximation of the posterior $Q_\theta(w)$, one can instead estimate the model-dependent uncertainty by the sample variance of $\theta$ (or the sample variance of $w$ in the case of non-Bayesian networks). To be precise, sampling mini-batches $\mathcal{B}_i$ from a batch $\mathcal{B}$, one can compute the adapted parameters by performing one gradient step (at $\theta$): $\theta'_i=\theta-\eta\nabla_\theta\mathcal{L}_{\mathcal{B}_i}(f_\theta)$, where $\mathcal{L}$ is the objective and $\eta$ is the inner learning-rate. Then the variance of $\theta$ can be defined as the trace of the covariance of vectorized $\theta'_i$s: 
\begin{align}
    \text{Var}(\{\theta'_i\})=\text{trace}(\text{Cov}({\{\text{vec}(\theta'_i)\}})),
    \label{eq:wvar}
\end{align}
\noindent where $\text{Cov}(\{\text{vec}(\theta'_i)\})=\eta^2 \text{Cov}\left(\{\text{vec}(\nabla_\theta\mathcal{L}_{\mathcal{B}_i}(f_\theta))\}\right)$ and $\{\cdot\}$ denotes a collection or a set. It can be easily seen that regularizing the variance of parameters is essentially regularizing the variance of gradients. We will discuss the usage of this gradient variance as a regularizer as well as its relationship with MAML~\cite{finn2017model} in the next section.

\section{Domain Adaptation via Calibrating Uncertainties}
\noindent \textbf{R\'{e}nyi entropy regularization.} Denote source and target dataset as $\mathcal{D}_S=\{x^{(s)}, y^{(s)}\}_{s\in \mathcal{S}}$ and $\mathcal{D}_T=\{x^{(t)}\}_{t\in \mathcal{T}}$ respectively, where $x^{s},x^{t}$ indicate the samples and $y^{s}$ is the label in source domain, and $\mathcal{D}=\mathcal{D}_S\cup\mathcal{D}_T$. We propose to calibrate the predictive uncertainty of target dataset with the source domain uncertainties. Concretely, we minimize the cross-entropy (CE) loss in the source domain while constraining the predicted entropy in the target domain:
\begin{align}
    \min_{\theta} \mathcal{L}_{CE}&=\frac{1}{|\mathcal{S}|}\sum_{s\in \mathcal{S}}{\text{H}_{CE}(y^{(s)}, P(y|x^{(s)}; \theta))} \nonumber \\
    \text{s.t.}\quad &\frac{1}{|\mathcal{T}|}\sum_{t\in \mathcal{T}}{\text{H}_\alpha(P(y|x^{(t)}; \theta))} \leq C,
    \label{eq:ce}
\end{align}
where $\text{H}_{CE}(\cdot,\cdot)$ is the cross-entropy and $C$ indicates the strength of the applied constraint. In practice, the network is first pretrained on labeled source dataset using ELBO in Equation~\ref{eq:elbo}. Then, unlabeled target data is introduced in the above Equation~\ref{eq:ce}, and $P(y|x;\theta)$ is computed from Equation~\ref{eq:pred}. Note that the resulting CE loss is no longer the term (I) in ELBO, since the expectation is inside logarithm. We simply treat $Q_\theta(w)$ ``as is'' the true posterior and evaluate CE using $P(y|x;\theta)$. For a non-Bayesian network, $P(y|x,w)$ is used as a replacement of $P(y|x;\theta)$.

To solve Equation~\ref{eq:ce}, rewrite it as a Lagrangian with a multiplier $\beta$,
\begin{align}
    \mathcal{F}=&\frac{1}{|\mathcal{S}|}\sum_{s\in \mathcal{S}}{\text{H}_{CE}(y^{(s)}, P(y|x^{(s)}; \theta))} + \nonumber \\
    &\beta \left( \frac{1}{|\mathcal{T}|}\sum_{t\in \mathcal{T}}{\text{H}_\alpha(P(y|x^{(t)}; \theta))}-C \right).
    \label{eq:lag}
\end{align}
Since $\beta, C \geq 0$, an upper bound on $\mathcal{F}$ is obtained,
\begin{align}
    \mathcal{F}\leq &\frac{1}{|\mathcal{S}|}\sum_{s\in \mathcal{S}}{\text{H}_{CE}(y^{(s)}, P(y|x^{(s)}; \theta))} + \nonumber \\
    & \frac{\beta}{|\mathcal{T}|}\sum_{t\in \mathcal{T}}{\text{H}_\alpha(P(y|x^{(t)}; \theta))} = \mathcal{L}_{\alpha}.
    \label{eq:ent}
\end{align}
Ideally, Equation~\ref{eq:lag} can be optimized via dual gradient descent and $\beta$ is jointly updated along with $\theta$. For simplicity, we follow the work of~\cite{higgins2017beta} and fix $\beta$ as a hyper-parameter in the experiment and minimize the upper bound $\mathcal{L}_{\alpha}$. 

Note that letting $\alpha \rightarrow 1$ in Equation~\ref{eq:ent} is in fact the (Shannon) entropy regularization as described in~\cite{grandvalet2005semi,grandvalet2006entropy}, except that here we consider a variational BNN. As pointed out in~\cite{grandvalet2006entropy}, directly optimizing Equation~\ref{eq:ent} can be difficult and expectation maximization (EM) algorithms are often used. Proposed in~\cite{yuille1994statistical,grandvalet2006entropy}, deterministic annealing EM anneals the predicted probabilities as soft-labels and minimizes the resulting cross-entropy. In an extreme case, soft-labels become one-hot vectors and the algorithm terms out to be self-training with pseudo-labels~\cite{lee2013pseudo}. In our R\'{e}nyi entropy regularization framework, self-training is essentially optimizing the min-entropy ($\alpha \rightarrow \infty$). Then the objective reads
\begin{align}
    \mathcal{L}_{\infty}=&\frac{1}{|\mathcal{S}|}\sum_{s\in \mathcal{S}}{\text{H}_{CE}(y^{(s)}, P(y|x^{(s)}; \theta))} + \nonumber \\
    & \frac{\beta}{|\mathcal{T}|}\sum_{t\in \mathcal{T}}{\text{H}_{CE}(\hat{y}^{(t)}, P(y|x^{(t)}; \theta))},
    \label{eq:st}
\end{align}
with $\hat{y}^{(t)}=\text{onehot}(\argmax_{k\in\{1,\ldots,K\}}{P(y_k|x^{(t)};\theta)})$ to be pseudo-labels in target domain. Subscript $k$ denotes the $k$-th element in a given $K$-dim vector. The relationship between $\mathcal{L}_{1}$ and $\mathcal{L}_{\infty}$ can be immediately realized by noticing that the Shannon entropy is an upper bound of the min-entropy:
\begin{align}
    \text{H}_1(P)=&-\sum_k{P_k \log(P_k)} \geq -\sum_k{P_k} \log(\max_k{P_k})= \nonumber\\
    &-\log(\max_k{P_k}) =\text{H}_\infty(P) =\text{H}_{CE}(\hat{y}, P).
\end{align}

We build our method on top of class-balanced self-training (CBST) proposed in~\cite{zou2018unsupervised} and use it as the backbone of RER. CBST seeks to select most confident predictions pseudo-labels in a self-paced (``easy-to-hard'') scheme, since jointly learning the model and optimizing pseudo-labels on all unlabeled data is naturally difficult. The authors also propose to normalize the class-wise confidence levels in pseudo-label generation to balance the class distribution. For a detailed formulation, we suggest readers referring Section 4.1 and 4.2 in~\cite{zou2018unsupervised}.

\noindent \textbf{Gradient variance regularization.} The entropy regularization or self-training framework as formulated above implicitly encourages cross-domain feature alignment. However, pseudo-labels can be quite noisy even if BNN is employed to estimated their reliability. Trusting all selected pseudo-labels as one-hot-encoded ``ground-truth'' is overconfident and self-training with noisy pseudo-labels can lead to incorrect entropy minimization. Indeed, we observe that the model can quickly converge to its overconfident predictions. Therefore, the parameter variance evaluated in target domain using pseudo-labels via Equation~\ref{eq:wvar} can be even smaller than that of the source domain. To address this problem, we again propose to regularize the self-training by maximizing the gradient variance. Algorithm~\ref{alg:gvr} illustrates the regularized self-training procedure on target domain (the training on source and target domains are preformed alternately, which is omitted in the algorithm box). $\eta$ and $\eta'$ are the inner- and outer-stepsize, and $\lambda$ is the hyper-parameter weighting the regularization term.
\begin{algorithm}[h]
\caption{Gradient Variance Regularization}
\label{alg:gvr}
\begin{algorithmic}[1]
    \While{not done}
        \State Sample mini-batches $\mathcal{B}_i \sim \mathcal{D}_T$
        \For{all $\mathcal{B}_i$}
            \State Evaluate $\nabla_\theta\mathcal{L}_{\mathcal{B}_i}(f_\theta)$
            \State Compute $\theta'_i=\theta-\eta\nabla_\theta\mathcal{L}_{\mathcal{B}_i}(f_\theta)$
        \EndFor
        \State Collect $\Theta'=\{\theta'_i\}$
        \State Compute $\text{Var}(\Theta')=\text{trace}(\text{Cov}({\{\text{vec}(\theta'_i)\}}))$
        \State Update $\theta \leftarrow \theta - \eta'\nabla_\theta \left(\sum_i{\mathcal{L}_{\mathcal{B}_i}(f_\theta)}-\lambda \text{Var}(\Theta') \right)$
    \EndWhile
\end{algorithmic}
\end{algorithm}

Notice that the proposed GVR shares similarities with MAML~\cite{finn2017model}, comparing from a dynamical systems standpoint and despite that MAML samples mini-batches of different tasks. Taking a first-order Taylor expansion of the loss function around $\theta$,
\begin{align}
    \mathcal{L}_{\mathcal{B}_i}(f_{\theta'_i}) \approx \mathcal{L}_{\mathcal{B}_i}(f_{\theta})-\eta\Vert{\nabla_\theta\mathcal{L}_{\mathcal{B}_i}(f_\theta)}\Vert_2^2,
\end{align}
we demonstrate that MAML tries to maximize the sensitivity of the loss functions with respect to the parameters by maximizing the $L_2$ norms of the gradients. On the contrary, GVR maximizes the variance of gradients, which intuitively encourages the model to escape from bad local minima.

It is worth mentioning that GVR is not only model-agnostic but also objective-agnostic. This is useful when the regularizer itself is the objective to be optimized. Moreover, GVR is complementary to VAT~\cite{miyato2015distributional} since in VAT the gradient is computed with respect to input data. We conjecture that the data gradient somewhat captures the aleatoric (data-dependent) uncertainty, which we leave for future work.

\section{Experiments}
We first show results on three digit datasets MNIST~\cite{lecun1998gradient}, USPS and SVHN~\cite{netzer2011reading}, where we consider \textbf{MNIST$\rightarrow$USPS} and \textbf{SVHN$\rightarrow$MNIST}. Then we present preliminary results on a challenging benchmark: \textbf{VisDA17} (classification)~\cite{peng2018visda} which contains 12 classes. We follow the standard protocol in~\cite{peng2018visda,tzeng2017adversarial,sankaranarayanan2018generate}. Classification accuracies on source and target domains for base models are reported in Table~\ref{tab:base}. We use DTN~\cite{zhang2015deep} as our base model for MNIST$\rightarrow$USPS and SVHN$\rightarrow$MNIST. To implement its Bayesian variant (BDTN), we add another classifier to predict the logarithm of variance. 
\begin{table}[h]
\centering
\subfloat[{\bf MNIST}\vspace{-0.02in}]{
\hspace{-0.1in}
\scalebox{0.8}{
    \begin{tabular}{lcc}
    \hline
    Model   & Source & Target \\
    \hline
    DTN       & 100.00 & 83.94 \\
    BDTN-M1   & 100.00 & 83.78 \\
    BDTN-M5   & 100.00 & 86.83 \\
    BDTN-M10  & 100.00 & 86.28 \\
    BDTN-M20  & 100.00 & 86.78 \\
    BDTN-M100 & 100.00 & 87.06 \\
    \hline
    \end{tabular}}
}
\subfloat[{\bf SVHN}\vspace{-0.02in}]{
\scalebox{0.8}{
    \begin{tabular}{lcc}
    \hline
    Model   & Source & Target \\
    \hline
    DTN       & 97.42 & 72.91 \\
    BDTN-M1   & 95.91 & 65.51 \\
    BDTN-M5   & 99.16 & 71.12 \\
    BDTN-M10  & 99.42 & 71.38 \\
    BDTN-M20  & 99.50 & 73.64 \\
    BDTN-M100 & 99.33 & 74.91 \\
    \hline
    \end{tabular}}
}
\caption{Training base models on MNIST and SVHN. BDTN is a modified Bayesian DTN~\cite{zhang2015deep}, with different $M$ values (as defined in Equation \ref{eq:pred}). Classification accuracies in source and target domains are reported.}
\label{tab:base}
\end{table}

Domain adaptation results on digit datasets are shown in Table~\ref{tab:toy}. Our proposed R\'{e}nyi entropy regularization methods with non-Bayesian and Bayesian base models are listed as RERs and BRERs respectively. We see self-training with pseudo-labels ((B)RER-$\infty$) are in general more stable than directly minimizing the Shannon entropy ((B)RER-1). Also, adding GVR in (B)RER-$\infty$ improves the performance. However, we also observe that GVR is not helpful in (B)RER-1 settings.

\begin{table}[h]
\centering
\subfloat[{\bf MNIST$\rightarrow$USPS}\vspace{-0.02in}]{
\scalebox{0.8}{
    \begin{tabular}{lcc}
    \hline
    Model   & Target Acc (\%) & Acc Gain (\%) \\
    \hline
    Source-DTN        & 83.94 & - \\
    Source-BDTN       & 84.89 & - \\
    RER-1             & 91.57$\pm$0.13 & 7.63 \\
    RER-1-GVR         & 91.97$\pm$0.26 & 8.03 \\
    RER-$\infty$      & 93.57$\pm$0.30 & 9.63 \\
    RER-$\infty$-GVR  & 93.88$\pm$0.14 & \textbf{9.94} \\
    BRER-1            & 92.78$\pm$0.42 & 7.89 \\
    BRER-1-GVR        & 93.07$\pm$0.72 & 8.18 \\
    BRER-$\infty$     & 94.42$\pm$0.12 & 9.53 \\
    BRER-$\infty$-GVR & \textbf{94.53$\pm$0.23} & 9.64 \\
    \hline
    \end{tabular}}
}\par
\subfloat[{\bf SVHN$\rightarrow$MNIST}\vspace{-0.02in}]{
\scalebox{0.8}{
    \begin{tabular}{lcc}
    \hline
    Model   & Target Acc (\%) & Acc Gain (\%) \\
    \hline
    Source-DTN        & 64.48 & - \\
    Source-BDTN       & 70.98 & - \\
    RER-1             & 88.46$\pm$0.90 & 23.98 \\
    RER-1-GVR         & 85.48$\pm$4.71 & 21.00 \\
    RER-$\infty$      & 88.16$\pm$1.19 & 23.68 \\
    RER-$\infty$-GVR  & 90.31$\pm$2.31 & \textbf{25.83} \\
    BRER-1            & 92.49$\pm$4.73 & 21.51 \\
    BRER-1-GVR        & 92.37$\pm$4.76 & 21.39 \\
    BRER-$\infty$     & 96.06$\pm$0.68 & 25.08 \\
    BRER-$\infty$-GVR & \textbf{96.38$\pm$0.05} & 25.40 \\
    \hline
    \end{tabular}}
}
\caption{Results on MNIST$\rightarrow$USPS and SVHN$\rightarrow$MNIST. RER uses DTN~\cite{zhang2015deep} as the base model. BRER-$\infty$ uses BDTN as the base model and optimizes $\mathcal{L}_{\infty}$, while BRER-1 optimizes $\mathcal{L}_{1}$. Results are averaged over 4 runs with different random seeds.}
\label{tab:toy}
\end{table}

Mean accuracies on VisDA17 dataset are reported in Table~\ref{tab:visda}. Following the protocol in~\cite{zou2018unsupervised}, we train a standard ResNet101~\cite{he2016deep} as the base model and add a second classifier (denoted as BRes101) to predict the logarithm of variance on logits. Results show that BNN improves upon non-Bayesian baselines by a large margin. GVR has not been tested on VisDA17 with (B)Res101 since the memory requirement exceeds our GPU capacities.

\begin{table}[h]
    \centering
    \scalebox{0.8}{
    \begin{tabular}{lcc}
    \hline
    Model   & Target mean-Acc (\%) & Acc Gain (\%) \\
    \hline
    Source-Res101  & 48.02 & - \\
    Source-BRes101 & 46.03 & - \\
    MMD~\cite{long2015learning} & 61.1 & - \\
    GTA-Res152~\cite{sankaranarayanan2018generate} & 77.1 & - \\
    \hline
    RER-$\infty$  & 76.81$\pm$2.73 & 28.79 \\
    BRER-$\infty$ & \textbf{80.59$\pm$1.39} & \textbf{34.56} \\
    \hline
    \end{tabular}}
    \caption{Preliminary results on VisDA17~\cite{peng2018visda} classification benchmark (validation set). Results are averaged over 5 runs with different random seeds.}
    \label{tab:visda}
\end{table}

\section{Conclusion}
In this work, we propose to approach unsupervised domain adaptation via calibrating the predictive uncertainties between source and target domains. The uncertainty is quantified under a general R\'{e}nyi entropy regularization framework, within which we introduce Bayesian neural networks for accurate and reliable uncertainty estimations. From a frequentist point of view, we in addition propose to approximate the model uncertainty via the sample variance of network parameters (or gradients) during training. Results show that the uncertainty estimation by Bayesian networks and gradient variances is effective and leads to stable performance in unsupervised domain adaptation.

{\small
\bibliographystyle{ieee}
\bibliography{egbib}
}

\end{document}